\documentclass[submit]{elsarticle}

\usepackage{hyperref}
\usepackage{xcolor}
\usepackage{amsmath}
\usepackage{times}
\usepackage{soul}
\usepackage{epsfig}
\usepackage{amssymb}
\usepackage{subcaption} 
\usepackage{multirow}
\usepackage{booktabs}
\usepackage{contour}
\usepackage{ulem}

\contourlength{0.8pt}

\begin{document}

\begin{frontmatter}

\title{3D Pose Estimation and Future Motion Prediction from 2D Images}

\author[add1]{Ji Yang}
\author[add1]{Youdong Ma}
\author[add1]{Xinxin Zuo}
\author[add1]{Sen Wang}
\author[add2]{Minglun Gong}
\author[add1]{Li Cheng\corref{corrauthor}}

\address[add1]{Department of Electrical and Computer Engineering,
	University of Alberta, Edmonton, AB, Canada}
\address[add2]{School of Computer Science, University of Guelph, Guelph, ON, Canada}

\cortext[corrauthor]{Corresponding author}


\begin{abstract}
This paper considers to jointly tackle the highly correlated tasks of estimating 3D human body poses and predicting future 3D motions from RGB image sequences. 
Based on Lie algebra pose representation, a novel self-projection mechanism is proposed that naturally preserves human motion kinematics. 
This is further facilitated by a sequence-to-sequence multi-task architecture based on an encoder-decoder topology, 
which enables us to tap into the common ground shared by both tasks.
Finally, a global refinement module is proposed to boost the performance of our framework.
The effectiveness of our approach, called PoseMoNet, is demonstrated by ablation tests and empirical evaluations on Human3.6M and HumanEva-I benchmark, where competitive performance is obtained comparing to the state-of-the-arts.
\end{abstract}


\end{frontmatter}

\section{Introduction}
\label{sec:intro}

Obtaining three-dimensional human poses from monocular image data is a fundamental yet challenging problem in computer vision. In many recent efforts~\cite{SemGCN,SimpleYet,2DMatching,MonocularNie17}, 3D human pose estimation has been decomposed into a two-stage process: first, the 2D keypoints that correspond to the body joints are detected from the 2D image, after which the detected joints are lifted to obtain 3D pose. 
This type of solution is elegant in terms of the simplicity of problem formulation, unfortunately it suffers from inherent ambiguities caused by projection: 
different 3D poses can share the same 2D pose projection given a specific viewpoint; that is, the mapping between the 2D joints detection and 3D pose is not bijective.
To resolve this ambiguity of 3D pose estimation from a monocular image, video-based pose estimation is also investigated in the literature~\cite{VideoPose3D, ExploitingTemporal}. 
Existing video-based pose estimation methods, however, either need to observe a relatively long history (243 frames~\cite{VideoPose3D} or can only handle a short video sequence (4-6 frames~\cite{ExploitingTemporal}) to achieve their best results.

In the meantime, there has been a growing stream of research endeavours on the closely related problem of future motion prediction: 
by observing a past sequence of 3D poses, the future 3D motion could be generated~\cite{LongTermMotion,AdvGeoAware,Shuang,ERD}. 
However, the requirement of past 3D pose sequence could be problematic: the MoCap data can be obtained relatively easily in a controlled lab environment but it is nearly infeasible to be captured in the wild at a reasonable cost. 
Besides, in predicting the future motion, motion prediction takes as input a short history of 3D pose sequence, which, in a sense, could be considered as a generalization of video-based pose estimation, where instead of current poses, the output here is future poses.

\begin{figure}[t]
    \begin{center}
        \includegraphics[width=0.85\linewidth]{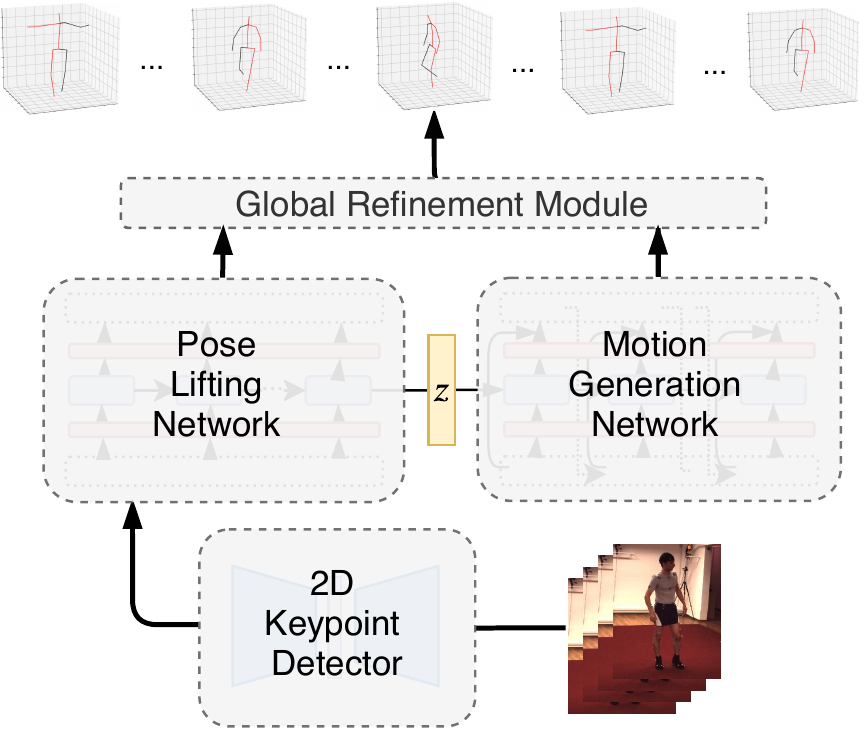}
    \end{center}
    \caption{The overview pipeline for our approach, PoseMoNet. The input to the pipeline is a sequence of observed video frames. A 2D keypoint detector is used to generate the corresponding 2D pose sequence. The predicted 2D pose sequence is then lifted to its 3D pose representation by the Pose Lifting Network (PLN). The Motion Generation Network uses the latent hidden state (from PLN) and the last estimated 3D pose as the seed pose to generate future motion. Finally, the entire predicted past pose and future motion sequence is refined together by the global refinement module.}
    \label{fig:overview}
\end{figure}

Motivated by these observations, in this paper, we propose an approach to jointly address the closely related problems of inferring 3D human poses and predicting future motions from an RGB video.
It not only allows us to tap into the common ground of the human motion dynamics shared by both problems but also facilitates the removal of the unnecessary MoCap constraints from the problem of future 3D motion prediction. 
A sequence-to-sequence multi-task approach is proposed to exploit the benefits of pose estimation from RGB video and future motion prediction from a partial RGB image sequence. 
A simple yet effective encoder-decoder framework back-boned by Recurrent Neural Networks (RNNs) called PoseMoNet is thus proposed to tackle both problems. As depicted in Fig.~\ref{fig:overview}, it contains four major components:
(1) a pre-trained 2D keypoint detector is used to detect the 2D pose of a human for each of the input video frame; 
(2) a Pose Lifting Network (PLN) maps a 2D pose sequence to 3D pose sequence and encodes the motion dynamics in a latent representation; 
(3) the latent representation and a predicted seed pose from the last observed time step are passed to the Motion Generation Network (MGN) to predict future motions;  
and (4) a global trajectory refinement module will further polish the entire trajectory estimated by both the PLN and the MGN. 
While in existing encoder-decoder architectures for motion prediction \cite{AdvGeoAware,Shuang,Seq2seq-ZeroV,ERD}, the role of the encoder is to extract temporal motion dynamic from past 3D pose sequence to a latent representation and the decoder will then derive the future motion from it, our PLN encoder only requires 2D pose sequence as input. 
In short, our PLN tries to estimate the corresponding 3D pose sequence from the given 2D pose sequence as well as extract and compress the motion dynamics.
It is worth mentioning that although the deep learning methods can generate future dynamics and achieve decent results on quantitative metrics, studies \cite{Shuang,Seq2seq-ZeroV} have shown the generated future motion tends to converge to a mean pose. 

A stronger evidence\cite{Seq2seq-ZeroV} reveals that, quantitatively, predicting future motion as with a zero-velocity baseline can beat many well-trained deep learning methods.
In other words, the trained model, instead of being trained to really understand human dynamics and predict the future motion, is actually overfitted to the quantitative metric.
We address this issue by adopting a Lie algebra-based pose representation and propose a self-projection mechanism that encourages the generated future motion to be more realistic and dynamic.

Our work is motivated and inspired by existing literature \cite{SimpleYet,Shuang,Seq2seq-ZeroV,ERD} but different from them. 
First, none of the current literature attempts to handle 3D pose sequence estimation and future motion prediction simultaneously. 
We utilize this multitask setting and shows that by pairing future motion prediction with 3D pose sequence estimation, the learning system gains robustness and achieves better performance with smaller joints error on both tasks. 
For the future motion prediction task, our proposed framework eliminates the requirement of the ground truth past 3D pose sequence and the input of a ground truth seed pose, where only the 2D video sequence is leveraged instead.
Furthermore, our framework shows that it is unnecessary to involve past 3D pose sequences to generate the future as it can be achieved from 2D pose sequence input by our pose estimation task.

Our main contributions are summarized below:
\begin{itemize}
    \item To our best knowledge, this is the first work to jointly tackle the closely related two problems of 3D pose sequence estimation and future motion prediction. 
    Given a short video sequence as an initial part of a motion sequence, 
    our approach, PoseMoNet, estimates not only the poses of these existing frames but also those of future frames to complete the entire motion dynamics. 
    \item 
    Based on Lie algebra pose representation, a sequence-to-sequence multi-task framework is proposed to address this relatively new problem. 
    Empirically, it is demonstrated to be capable of predicting future motion dynamics faithfully, as well as maintaining a competitive estimation of 3D poses from a partial sequence of RGB images.
\end{itemize}

\section{Related Work}
\label{sec:related_work}

\subsection{3D Pose Estimation}

3D human pose estimation research is affected by deep learning significantly in recent years where conventional methods \cite{Pictorial, Graphical} are overtaken by deep learning methods.
As the 2D human pose estimation results are progressively improved, researchers have also started to use detected 2D keypoints as an intermediate for 3D human pose estimation. 
Many works try to build algorithms that map the 2D keypoints to a 3D pose, such as using the nearest neighbor search~\cite{2DMatching}, or building a parametrized method with neural networks to approximate it~\cite{SimpleYet}. However, body parts occlusion is a critical issue for these types of 2D-to-3D methods, \cite{Occlusion} proposed to use multi-scale heatmaps together with an occlusion augmentation module to further resolve the occlusion problem. 
The graphical structure of human pose is also leveraged by novel graph convolutional network methods. 
In \cite{pr1_tian2021adversarial}, the authors further inject the graph structure into a GAN-based human pose estimation model to capture the relationship of body joints. Pose Graph
Convolutional Network (PGCN) \cite{pr2_bin2020structure} is introduced to exploit structural relationships between body key points to further improve the localization performance for pose estimation. 
A similar idea is examined from a different perspective in \cite{review5_2_lee2018propagating}, joint interdependency is leveraged to estimate 3D joint position from RGB images by incorporating structural relationships of the human body, and a progressively designed propagating LSTM framework is used to tackle the pose estimation stage by stage.
In \cite{review5_3_cai2018weakly, review5_1_cai20203d}, weakly-supervised methods are proposed to leverage both synthetic data and weakly-labeled real-world data, where the depth-aid regularizer provides weak supervision for 3D pose estimation without any costly 3D annotation required.
Further improvements are investigated by introducing the temporal dimension at the input. 
A sequence of 2D keypoints are used to provide smoother spatial constraints where~\cite{ExploitingTemporal} leverages recurrent models and~\cite{VideoPose3D} built a 1D convolution model with optional temporal dilation. 
A grounded spatial-temporal learning framework was proposed in~\cite{ST-GCN-Pose} to leverage both the temporal context in the video sequence and the spatial information in the graph-based skeleton. In the light of exploring spatial-temporal learning, \cite{Anatomy} use the entire video as the context for predicting the bone direction along with a consistent bone length across the entire video. \cite{xu2020deep} proposed a multi-step refinement and estimation framework that refines the 2D input keypoint sequence and then concurrently considering the structure of 2D inputs and 3D outputs.
Interestingly, another family of methods \cite{Learning,Predicting, VIBE} train deep models to directly model the human shape represented by SMPL \cite{SMPL} and perform reverse inference from the shape to the corresponding pose.

\subsection{Human Motion Prediction}

The essence of motion prediction or generation can be thought of as a sequence generation task.
Conventional approaches leveraged Gaussian processes~\cite{GaussianMotion} and hidden Markov models~\cite{MarkovMotion} to learn the representation of human motion. 
With the recent release of several large-scale motion capture (MoCap) datasets~\cite{HumanEva,H36M}, most of the recent research efforts involved deep models, more specifically, recurrent neural networks. 
Encoder-Recurrent-Decoder (ERD) \cite{ERD} can be considered as one of the earliest attempts to leverage the superior learning power with deep models for human motion prediction. 
A residual Gated Rectified Unit network for generating future human dynamics is proposed in~\cite{Seq2seq-ZeroV}, as well as a zero-velocity baseline that outperforms previous works. Interestingly, the zero-velocity baseline outperforms \cite{ERD} by simply predicting the last observed pose as the result.
Structural-RNN~\cite{S-RNN} considers the body parts of humans and the object in the scene from a graph structure perspective though it requires human intervention to design such graph structures. 
Generative adversarial networks are employed and both the geometric property of articulated objects and adversarial learning concept are utilized to further push the quantitative results for predicting long-term human motion~\cite{AdvGeoAware}. The only drawback is the involvement of the classifier (discriminator) make the training process more sensitive and complicated.
Instead of using a recurrent architecture to model the motion context, a modified highway unit is utilized in~\cite{MHU}. 
And a hierarchical motion context network combined with a Lie algebra pose representation similar to~\cite{LieX} is introduced in~\cite{Shuang}. Furthermore, human dynamics can be learned by working directly in the trajectory space, instead of the traditionally used pose space \cite{TrajLearn}, by using discrete cosine transform. 
With recent advances of the attention mechanism, in \cite{mao2020history},  an attention-based feed-forward network is proposed to accommodate the observation that human motion tends to repeat itself. \cite{cai2020learning} applies a transformer-based architecture with the global attention mechanism as well as proposing a memory-based dictionary which help to preserve the global motion pattern.

\subsection{Multitask Learning}
Popular deep learning methods are usually built for a single particular task. However, potential benefits may be obtained through a multitask learning (MTL) setting on related tasks. Early works have shown that MTL is intuitively plausible to obtain inductive bias by including implicit data augmentation, regularization, and representation bias \cite{MTLearning}. Common MTL methods with deep learning usually follow different patterns, such as joint learning (parameter sharing) and learning with auxiliary tasks. Parameter sharing usually has two forms: hard sharing and soft sharing. Hard sharing comes with the form of using task-specific output layers while sharing the layers before task-specific layers among all tasks \cite{MTLearning}. Soft sharing usually means most of the layers in the neural networks are jointly trained together under the supervision from all tasks \cite{Low}. 

In the field of computer vision, significant efforts are put into leveraging MTL or joint learning.
An Identity-Aware CycleGAN (IACycleGAN) model that applies a new perceptual loss to supervise the image generation network is introduced for both face photo-sketch synthesis and recognition \cite{pr4_fang2020identity}. 
It is reasonable to improve the super-resolution performance on both the color and the depth image by incorporate mutual information shared by them together with a GAN-based architectural design \cite{pr3_zhao2019simultaneous}.
A 2D/3D multitask system is proposed in \cite{MTL2D3D} where uses the pose estimation part as the primary tasks and jointly leverage the learned features for fine-grained action recognition.
The idea of MTL seems to be effectively involved in existing works with an encoder-decoder pattern for different types of computer vision tasks.
Face alignment in the video is tackled with a recurrent encoder-decoder architecture in \cite{r1b_peng2016recurrent}, where the proposed framework predicts 2D facial keypoint maps and also a constrained shape response map. In \cite{r1a_zhao2018learning}, the authors proposed a multi-purpose framework for image-to-video translation, and a recurrent encoder-decoder architecture is used for face alignment as well as generating future motion masks. These works also motivate us to explore the possibility of applying this pattern for 3D human pose estimation though they are not performing the 3D human pose estimation task itself.

In particular, MTL also motivates the literature of pose estimation. A pose grammar mechanism that encodes kinematics, symmetry, and motor coordination, is proposed to tackle the problem of 3D human pose estimation \cite{PoseGrammar}. Jointly learning a head pose estimation network by leveraging a coarse-to-fine network is efficient and the synthetic data is involved in training \cite{pr5_WANG2019196}. Body part segmentation can also provide extra information for accurately estimated the pose consider that joints should not be predicted outside the body part and particular joints can only be located in specific body parts \cite{xia2017joint}. 
\section{Methodology}
\label{sec:method}

An overview of the proposed architecture is shown in Figure \ref{fig:overview}. There are 4 major components in our framework: (1) a 2D keypoint detector network, where we directly incorporate the stacked hourglass model \cite{Hourglass} to acquire 2D pose sequence as input to the encoder from video frames. 
(2) the encoder: a Pose Lifting Network (PLN) for 3D pose sequence estimation;
(3) the decoder: a Motion Generation Network (MGN) for predicting future motion;
(4) a Global Refinement (GR) module for further polishing the entire pose trajectory which contains the estimated 3D pose sequence and future motion sequence. In addition, both the PLN and the MGN contain a novel Self-Projection (SP) layer.

The input to the framework is a sequence of RGB video frames, denoted as $\langle I_{(1)}, ..., I_{(T)}\rangle$. These observed video frames are then fed to a 2D pose detector network to obtain a sequence of 2D keypoints in the form of normalized image coordinate. 
In this work, we use a 16-joint human skeleton model.
The 2D pose sequence is then passed to the Pose Lifting Network (PLN) that lifts the 2D pose sequence to its predicted corresponding 3D representations.
A Motion Generation Network (MGN) that uses the last predicted 3D pose as the seed pose and the latent representation from the PLN to predict future poses. 
Finally, a global refinement (GR) module is applied over the entire predicted sequence, including both the predicted 3D pose and the 3D future motion.
Furthermore, we consider the commonly used 3D coordinate representation as a good baseline though it suffers from the problem of lacking physical constraints. In addition, a Lie algebra representation is used as to provide more grounded physical constraints, especially the 3D ground truth past sequence is not used in our work as the common 3D future motion prediction. Therefore, at each time step, two 3D poses are estimated (applied for both PLN and MGN).

In the rest of this section, we first introduce two pose representations we leveraged throughout our work, followed by a detailed demonstration of the architecture of our approach, PoseMoNet, as shown in Figure \ref{fig:encoder-decoder}. Finally, we conclude this section with a brief analysis of the multitask foundation of our work.

\subsection{Pose Representation}

\begin{figure}[t]
	\begin{center}
		\includegraphics[width=0.55\linewidth]{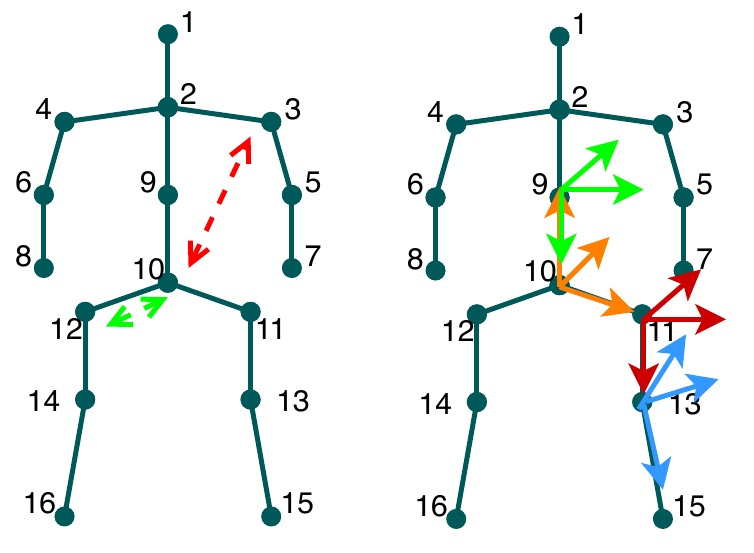}
	\end{center}
	\caption{Left: a relative coordinate-based pose, joint 10 (hip center/pelvis) is used as the root joint. All the other joints is represented by a relative position to the root joint. Right: the Lie algebra-based pose, while each bone itself maintains a local coordinate system, the position of each bone except for the one selected as root bone (joint 9-10) can be inferred by the nature of a kinematic chain by using ordinary matrix transformation. There are 5 kinematic chains used in the Lie-algebra-based pose, 3 of them are rooted from joint 10, namely the main back and two legs. The two chains represent two arms are originated from joint 2 and therefore can be also inferred from joint 10 with rotation angles.}
	\label{fig:skeleton}
\end{figure}

The coordinate-based pose representation is straightforward. 
It is the relative position of all joints with respect to the hip joint under a specified camera coordinate system, as shown in Figure~\ref{fig:skeleton} (left). 
However, this type of representation contains no physical constraints and no other information such as rotation except for the spatial position of joints.
If this is unclear to you, imagine to rotate your wrist and you should be able to notice that although those are different poses, the relative position of your wrist to other joints remains the same as before.

Based on the above observation, we introduce a Lie algebra pose representation that utilizes the theory of Lie groups~\cite{LieX,Shuang}. 
The human skeleton structure makes humans articulated objects, which can be easily characterized as a kinematic tree of rigid bones connected by joints. 
A kinematic chain forms a model for the skeletal structure, which can be considered as an assembly of bones connected and constrained by the bones.
The relative geometry of the successive bones, $b_{i}$ and $b_{i+1}$, can be represented by a point in the Special Euclidean group $SE(3)$. 

We use the example shown in Figure~\ref{fig:skeleton} (right) to illustrate the setup. 
For the bone contains two joints, i and j, denoted as $b_{i:j}$, a local coordinate system is set to use joint i as the origin and the $x$-axis aligned to the bone in the direction from i to j. 
Then a 3D rigid transformation can be used to transform the local coordinate system at bone $b_{i:j}$ to the one that represents the successive bone $b_{j:h}$. 
To summary, the 3D rigid transformation, which is an element of $SE(3)$, is denoted as a $4\times4$ matrix of the form
$\left(\begin{matrix}\mathrm{R} & \mathrm{\mathbf{t}} \\ 0 & 1\end{matrix}\right)$, 
with $\mathrm{R}$ being a $3\times3$ rotation matrix, and $\mathrm{\mathbf{t}}$ a 3D translation vector. 
Mathematically, the joint with coordinates $\mathrm{\mathbf{x}} = (x,y,z)^\intercal$ w.r.t. coordinate system at $b_{\{j:h\}}$ will have coordinates $\mathrm{\mathbf{x}}' = (x',y',z')^\intercal$ w.r.t coordinate system $b_{\{i:j\}}$ with

\begin{equation*}
\left(\begin{matrix}
\mathrm{\mathbf{x}}' \\ 1
\end{matrix} \right)
=
\left(\begin{matrix}
\mathrm{R}_{i:j} & \mathrm{\mathbf{t}}_{i:j} \\
0 & 1
\end{matrix} \right)
\left(\begin{matrix}
\mathrm{\mathbf{x}} \\ 1
\end{matrix} \right)
\end{equation*}

Therefore, the entire forward chain is naturally represented as the product of a ground of 3D rigid transformations and an entire human skeleton can be constructed by several such chains. 
The Lie group manifold contains motions as curves, however, it is non-trivial to parametrize and regress curves on this manifold \cite{Invitation}. 

In other words, Lie groups are geometric objects, i.e., manifolds, while Lie algebras are linear objects, i.e., vector spaces. Therefore, it can beneficial to use Lie algebra parameters over the SE(3) matrix representation from the computation perspective.
The product that represents a kinematic chain in $ SE(3) \times SE(3) \times \cdots \times SE(3)$ needs to be mapped to its Lie algebra space, denoted as $SE(3)\to \mathfrak{se}(3)$.

Lie algebra $\mathfrak{se}(3)$ refers to the identity of $SE(3)$ in the tangent space \cite{Invitation}. A logarithm map can be used to associate $SE(3)\to \mathfrak{se}(3)$,
\begin{equation}
\log: \left(\begin{matrix}\mathrm{R} & \mathrm{\mathbf{t}} \\ 0 & 1\end{matrix}\right) \mapsto\xi_\times 
\end{equation}
where $\xi_\times$ can be comfortably mapped to the vector form,
$\xi=(\mathbf{\omega}, \mathbf{\nu})^T$.
Readers may refer to \cite{Invitation} for a closed-form solution and its mathematical proof and deviation details. 

The above builds the foundation of recasting a skeletal pose as a $\mathfrak{se}(3)$ parameterized vector, $\mathrm{\mathbf{p^{\text{Lie}}}}=\left({\xi_{1}^{1}}^\intercal,\cdots,{\xi^{1}_{m_1}}^\intercal,\cdots,{\xi^{K}_{1}}^\intercal,\cdots,{\xi_{m_K}^{K}}^\intercal\right)^\intercal$.
$K$ denotes the number of kinematic chains (which is 5 in our case), $m_k$ is the number of joints in the $k$-th chain, and $\xi^k_i$ the Lie algebra parameter vector of joint $i$ in chain $k$. We can again look at the right plot in Figure \ref{fig:skeleton}, 

\subsection{PoseMoNet}

In this section, we introduce the details of the architectural design of each component in the PoseMoNet by first presenting a novel Self-Projection module. Then we demonstrate the Pose Lifting Network (PLN) followed by the Motion Generator Network (MGN). The Global Refinement (GR) module is used to further refine the estimated result with a grid-based pose representation.

\subsubsection*{The Self-Projection Module}
\begin{figure}[t]
	\begin{center}
		\includegraphics[width=0.8\linewidth]{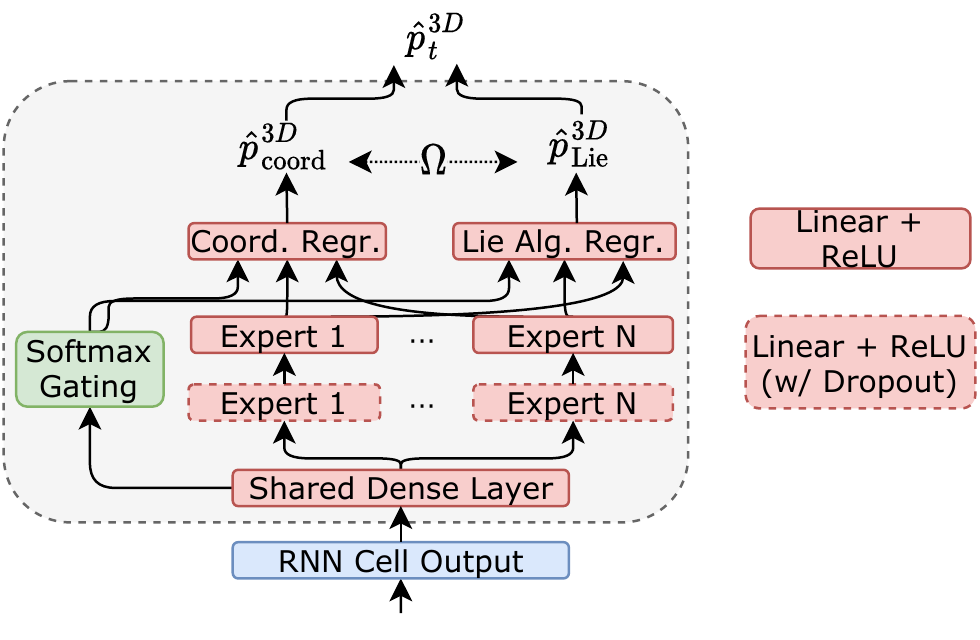}
	\end{center}
	\vspace{-0.5em}
	\caption{The self-projection module based on a mixture-of-expert mechanism. Note that due to the limited space, we only draw two expert paths but there can be $N$ such paths where we use 5 experts. The module also behaves as the output layer at each time step. It predicts the two types of pose representation and the regularizer $\Omega$ is the L2-norm of the difference between the two pose representations after forward kinematic inference of the Lie algebra pose representation $\hat{p}^{\text{3D}}_{\text{Lie}}$.}
	\label{fig:self_proj}
\end{figure}

Previous works investigated the usage of Lie algebra-based pose representation \cite{Shuang,LieX} to provide implicit physical constraints for estimating the 3D pose of an articulated object. We further attempt to use the Lie algebra pose representation to consolidated the prediction. The self-projection module, which is a derived version of the Mixture-of-Expert layer for regressing both the joint relative coordinates and the Lie algebra pose representation, is shown in \ref{fig:self_proj}. A Mixture-of-Expert design of soft parameter sharing is utilized. The output from the RNN cell is the input of a shared layer followed by the two separated expert paths by using softmax gating.

Given a kinematic chain of m joints with the corresponding parametrized Lie algebra pose $\mathrm{\mathbf{p^{\text{Lie}}}}=\left({\xi_{1}^{1}}^\intercal,\cdots,{\xi^{1}_{m_1}}^\intercal,\cdots,{\xi^{K}_{1}}^\intercal,\cdots,{\xi_{m_K}^{K}}^\intercal\right)^\intercal$, we are able to obtain the location of the Joint $\mathbf{J}_i$ by a completely differentiable forward kinematics
\begin{equation*}
\left(\begin{matrix}{\mathbf{J}}_i \\ 1 \end{matrix}\right) = \left[\prod_{j=1}^{i}\exp(\xi_{j\times})\right] \left(\begin{matrix} \mathbf{0} \\ 1 \end{matrix}\right).
\end{equation*}
At time step $t$, the desired output from both the coordinate pose path and the Lie algebra path should be well-aligned after the forward kinematic inference if the predictions are perfect. The training process should be able to penalize the bad prediction where the two outputs are very different after the kinematic chain forwarding. We define this process as $\mathbf{p}^{\text{Coord}}_{(t)} = f_{\text{kin}}(\mathbf{p}^{\text{Lie}}_{(t)})$. Note that this kinematic chain forwarding process contains inevitable tiny biases during the calculation, therefore we consider this as a regularization step for a more stable training process.
The L2 norm of the outputs of these two paths is then used as a regularization term for supervision, denoted as 
$$\Omega_{(t)} = \| \mathbf{p}^{\text{3D}}_{(t)} - f_{\text{kin}}(\mathbf{p}^{\text{Lie}}_{(t)}) \|_2$$

\subsubsection*{Pose Lifting Network}

\begin{figure*}[t]
	\begin{center}
		\includegraphics[width=1.02\linewidth]{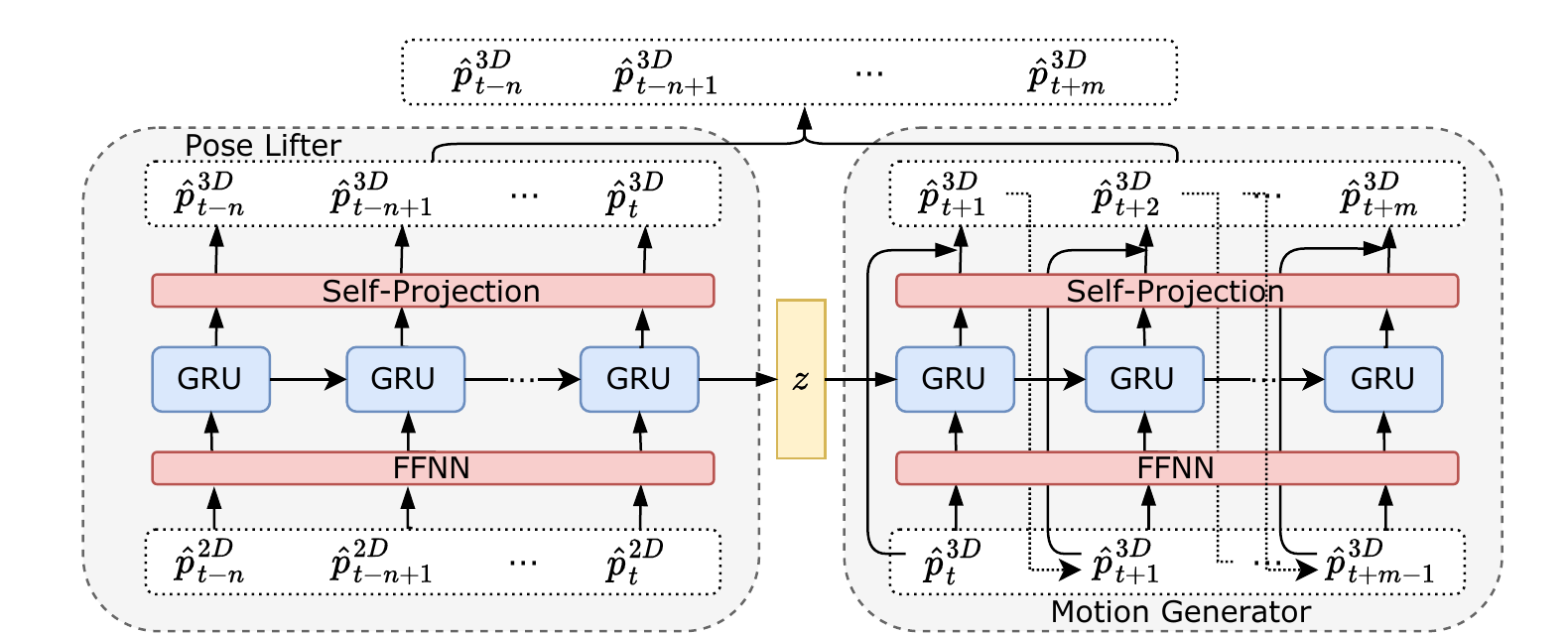}
	\end{center}
	\caption{A detailed illustration of the Pose Liter Network (PLN) and the Motion Generation Network (MGN) under the classic encoder-decoder architecture. PLN estimates the 3D pose sequence aligned with the time step of the input 2D pose sequence as well as encodes a latent representation for the decoder. MGN uses the 3D pose estimated at time step $t$ as the seed pose along with the latent representation encoded by PLN to generate future motion.}
	\label{fig:encoder-decoder}
\end{figure*}

The Pose Lifting Network (PLN) is designed to handle the task of lifting a 2D pose sequence to its corresponding 3D pose trajectory on its own duty, as well as behave as the encoder network in the overall framework. 

As shown in Figure \ref{fig:encoder-decoder}, the architecture design of the PLN is summarized as the following:
it consists of a feed-forward neural network (FFNN), then a 2-layer bi-directional Gated Rectified Unit (GRU), and finally a Self-Projection (SP) layer for producing the 3D pose trajectory. 
There are two linear layers in the FFNN and both are followed by a Rectified Linear Unit (ReLU) as the non-linearity. 
Dropout is added after all non-linearity layers except for the output layer from the SP layer, which does not have an activation function.

More formally, we define the input of the PLN as a length $T$ sequence of 2D pose $\mathbf{\hat{P}}^{\text{2D}_{(1:T)}} = \langle\mathbf{p}^{2D}_{(1)}, ... \mathbf{p}^{2D}_{(T)}\rangle$. 
There are two outputs from PLN: a latent representation of the entire sequence, $z_T$, and a predicted 3D pose sequence $\mathbf{\hat{P}}^{\text{3D}}_{(1:T)}$.
We note that $\mathbf{\hat{P}}^{\text{3D}}_{(1:T)} = \langle\mathbf{\hat{p}}^{3D}_{(1)}, ... \mathbf{\hat{p}}^{3D}_{(T)}\rangle$ where at each time step $t$, $\mathbf{\hat{p}}^{\text{3D}}_{(t)} = \langle\mathbf{\hat{p}}^{\text{Coord}}_{(t)}$, $\mathbf{\hat{p}}^{\text{Lie}}_{(t)}\rangle$ with
$\mathbf{\hat{p}}^{\text{Coord}}_{(t)} \in \mathbf{R}^{16 \times 3}$ and $\mathbf{\hat{p}}^{\text{Coord}}_{(t)} \in \mathbf{R}^{16 \times 6}$. The latent representation, $z_T$, is a tensor of shape $(\text{RNN layer \#}  \times \text{Hidden Size} \times 2)$. With abbreviating the forward pass as a function mapping $\phi_{\text{PLN}}$, the loss function for PLN is $$L_{\text{PLN}} = \frac{1}{T} \sum_{t=1}^{T} \| \mathbf{p}^{\text{3D}}_{(t)} - \phi_{\text{PLN}}(\mathbf{p}^{\text{3D}}_{(t)}) \|_2$$


\subsubsection*{Motion Generator Network}

The overall architecture of the Motion Generator Network (MGN) is similar to the PLN with the following differences, as shown at the right side of Figure \ref{fig:encoder-decoder}. Starting from the first step of motion generation at time $T+1$, the input is the predicted 3D pose $\mathbf{\hat{p}}^{\text{3D}}_{(T)}$ instead of the ground truth 3D pose $\mathbf{p}^{\text{3D}}_{(T)}$. To our knowledge, this setup also differs from all previous works \cite{AdvGeoAware,Shuang,Seq2seq-ZeroV,ERD}. We follow the common practice presented in previous works \cite{Seq2seq-ZeroV} to predict the residual of the motion from the previous time step. 

Mathematically, at time step $T + 1$, the input of the MGN is $\mathbf{\hat{p}}^{\text{3D}}_{(T)}$ and the latent representation $z_T$. As before, we use a function mapping $\phi_{\text{MGN}}$ to abbreviate the forward pass in MGN. Consider the future motion prediction is a $K$-step process, for every time step $t$, where $T + 1 \leq t \leq T + K$, the predicted 3D pose $\mathbf{\hat{p}}^{\text{3D}}_{(t)} = \mathbf{\hat{p}}^{\text{3D}}_{(t - 1)} + f_{\text{MGN}}(\mathbf{\hat{p}}^{\text{3D}}_{(t - 1)}, z_{t-1})$. We shorten the entire step-by-step process as $\mathbf{\hat{P}}^{\text{3D}}_{T+1:T+M} = \phi_{\text{MGN}}(\mathbf{\hat{p}}^{\text{3D}}_{(T)}, z_{t-1})$. The loss function for the MGN is then $$L_{\text{MGN}} = \frac{1}{K} \sum_{t=T+1}^{T+K} \| \mathbf{p}^{\text{3D}}_{(t)} - \phi_{\text{MGN}}(\mathbf{\hat{p}}^{\text{3D}}_{(t-1)}, z_{t-1}) \|_2$$


\subsubsection*{Global Refinement Module}

\begin{figure}[t]
	\begin{center}
		\includegraphics[width=0.9\linewidth]{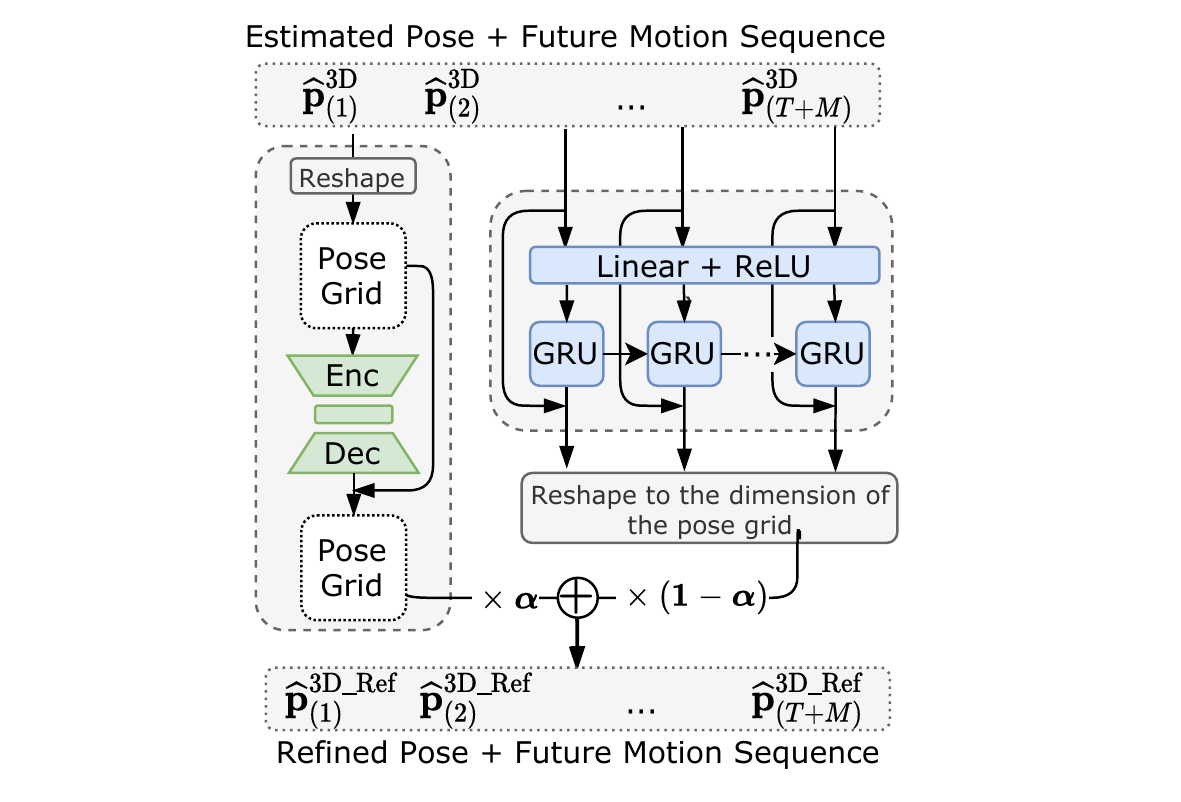}
	\end{center}
	\caption{An overview of the global refinement module. Left path: a residual convolutional encoder-decoder (ConvED) spatio-temporal refinement module. Right path: a simple GRU-based residual temporal refinement module (GRU-R). }
	\label{fig:refinement}
\end{figure}

We investigate two baseline architectures, the vanilla Gated Rectified Unit (GRU) and a Convolutional Encoder-Decoder (ConvED) to handle both the temporal and spatial context inside the pose trajectory, and then we propose a Global Refinement (GR) module built upon the two baselines for pose sequence refinement. 
The overall design in demonstrated in Figure \ref{fig:refinement}. 
Overall, the refinement is performed over the entire predicted pose trajectory, $\mathbf{\hat{P}}^{\text{3D}}_{(1:T+M)}$. 
The refinement module contains two paths and a trainable adaptive weighting vector, $\mathbf{\alpha} \in (0, 1)$. 
We denote the two forward passes that happened in the two paths as $\phi_{\text{GRU-R}}$ and $\phi_{\text{ConvED}}$. 
Formally, with $\odot$ refers to the Hadamard product, the forward pass of the GR Module can be represented as $\phi_{\text{GR}}(\mathbf{\hat{P}}^{\text{3D}}_{(1:T+M)}) = \mathbf{\alpha} \odot \phi_{\text{GRU-R}}(\mathbf{\hat{P}}^{\text{3D}}_{(1:T+M)}) + (\mathbf{1} - \mathbf{\alpha}) \odot \phi_{\text{ConvED}}(\mathbf{\hat{P}}^{\text{3D}}_{(1:T+M)})$, which gives us another loss function 
$$L_{\text{GR}} = 
\frac{1}{T + M} 
\sum_{t=1}^{T + M} 
\| \mathbf{p}^{\text{3D}}_{(t)} -
\phi_{\text{GR}}(\mathbf{\hat{p}}^{\text{3D}}_{(1:T+M)}) 
\|_2$$

\begin{figure}[t]
	\begin{center}
		\includegraphics[width=0.60\linewidth]{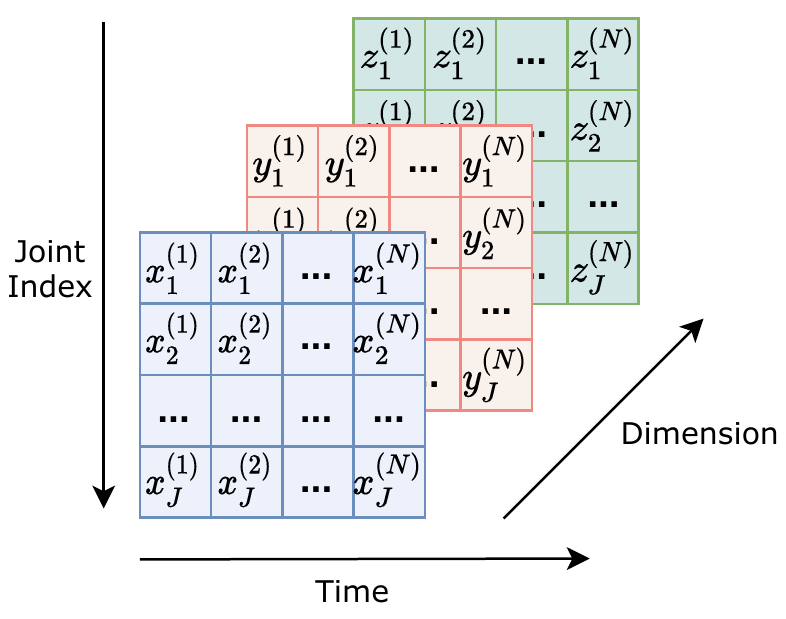}
	\end{center}
	\caption{A grid-based representation for pose trajectory. The pose trajectory has it shape as ($J \times N \times D$) which can be naturally considered as an image grid representation, where $J$ is the number of joints (height), $N$ is the temporal length of the trajectory (width), and $D$ is the dimension of the joint representation involved (3 for the coordinate-based and 6 for the Lie algebra-based representation).}
	\label{fig:pose_grid}
\end{figure}

The idea of using a 2D convolutional architecture to process pose trajectories can become intuitive if the pose trajectory is converted to a spatio-temporal grid representation which is identical to common digital images. 
A demonstration of the spatio-temporal pose grid representation is shown in Figure \ref{fig:pose_grid}.
Suppose the human skeleton model $J$ joints, and there are $N$ time steps and in total for the human pose trajectory. A built pose grid is then of shape $(J, N, D)$ where the height is the joint index dimension and the width is the temporal dimension. 
We only show the $D=3$, i.e., a 3-channel case to save the space but it can be easily generalized to a higher dimension case when the Lie Algebra representation is included (i.e., from 3 channels to 9 channels). 
While the adjacent sub-regions in the grid represent closely related joints, different sizes of receptive fields across different convolutional layers can, therefore, learn the spatial relationship of joints across the time dimension implicitly.

\subsubsection*{Loss Function}

The overall loss function which is a weighted combination of the loss functions defined in the above sections:
$$L = L_{\text{PLN}} + L_{\text{MGN}} + \beta L_{\text{GR}} + \lambda \Omega_{\text{SP}}.$$ In our experiments, we use $\beta = 0.2$ and $\lambda=0.01$. The value of the weights are chosen by empirical results. It is worth to notice that the $\Omega_{\text{SP}}$ term behaves more like a constraint noise regularizer as there are always small biases caused by the re-projection process, i.e., the process of converting the Lie algebra-pose to the coordinate-based pose. 
Using a weighted loss in our framework can be considered from two aspects. First, $\beta$ is used to control the balance of the importance between pose estimation and future motion prediction. Consider the stochasticity and a usually longer length of the temporal scope for motion generation, the weight of the motion generation loss term is reduced to 0.2. Second, the regularizer weight, $\lambda$, is used to re-scale the regularizer to a similar scale of the other terms in the loss function.

\subsection{Analysis of the Multitask Formulation}

A common practice for using an encoder-decoder or a sequence-to-sequence model for future motion prediction is that the encoder is only used for encoding the observed pose sequence \cite{AdvGeoAware,Shuang,Seq2seq-ZeroV,ERD} and it outputs a latent vector representation $z$ that ideally covers all the information of \textbf{the past 3D pose trajectory}. In our problem setup, instead, we use a past 2D pose sequence only. 

The question left is, does a 2D pose sequence contain sufficient information that represents the 3D dynamic of human motion? 
We build ablative experiments in Section \ref{sec:experiments} to test it and, unfortunately, the answer is negative. This observation drove the design of PLN of including a 3D pose sequence estimation task to force the model to encode sufficient information about the 3D dynamics from a 2D pose sequence in the latent representation. 

To sum up, the design of the overall framework reinforces the multitask foundation from two different granularity. First, with the incorporation of the Self-Projection layer, the outputs from recurrent cells are used to estimate two different types of pose representations where the soft parameter sharing mechanism contained in the SP layers can also introduce plausible regularization power and induction bias. It frames the multitask setting at a local level. Secondly, along with the MGN, the PLN is trained to not only lifting the 2D pose sequence to its corresponding 3D pose sequence but also try to estimate and encode 3D dynamics efficiently for generating reasonable future 3D motion, it concludes the multitask setup at the global level.

\section{Empirical Evaluations}
\label{sec:experiments}

\subsection{Dataset and Evaluation}
Human3.6M~\cite{H36M} is a large-scale dataset that contains 3.6 million video frames with 3D human poses captured by a MoCap system in an indoor environment. 
Eleven subjects who are professional actors perform 15 different daily activities such as eating, walking, smoking, discussing, and greeting. All of the performances are captured by 4 cameras placed at the pre-defined positions in the room so the intrinsic and extrinsic camera parameters are known. 
Only video frames from seven of the eleven subjects are annotated and publicly available, so we follow the previous work to split the training and testing set \cite{SemGCN,SimpleYet,MonocularNie17,VideoPose3D,ExploitingTemporal}. We downsample the video frame rate by a factor of 2 reducing the video from 50 MHz to 25 MHz. 
To align with the literature, we modify the skeleton data in Human3.6M to a 16-joint version. Note that we do not train specific models for each action but only train a multi-action model in an end-to-end fashion while early works may use multiple models and each for a single type of action only. 

HumanEva-I \cite{HumanEva}, on the other hand, is a relatively small dataset, where three subjects are recorded from three camera views at 60 Hz. Only three primary actions are used for evaluation, namely, Walk, Job and Boxing. Due to its limited size and reported issue with the corrupt frames \cite{VideoPose3D} which makes it is hard to be used for training and precise evaluation for pose estimation from video, we only show qualitative results on it.

For 3D pose sequence estimation, we adopt evaluation protocols from existing works \cite{VideoPose3D,ExploitingTemporal,SimpleYet,MonocularNie17,2DMatching}, namely Mean Per Joint Position Error (MPJPE) and Procrustes-MPJPE (P-MPJPE). Where MPJPE is the average Euclidean distance of predicted joints to the ground-truth and P-MPJPE is the MPJPE calculated after the estimated 3D pose is aligned to the ground-truth by the Procrustes method (a similarity transformation).
For future motion prediction, we benchmark the result with the most commonly used metric, Mean Angle Error (MAE) \cite{ERD,S-RNN,Seq2seq-ZeroV,Shuang,AdvGeoAware}. We also report the evaluation results by adopting the 2D ground truth pose sequence as input to demonstrate the lower bound of errors of our PoseMoNet.

\subsection{Implementation Detail}

For all recurrent models in the framework, the hidden state size is set to 512 with a 25\% dropout rate. 
We train the model with an initial learning rate of 0.001.
The training process lasts for a fixed number 30 epochs for Human3.6M with a batch size of 32 as well as decaying the learning rate every 10000 training steps by a factor of 0.9. Throughout the experiments, models are trained models with 9, 27, and 54 frames length of the input past 2D pose sequence along with a fixed 20-frame future motion context. As HumanEva-I is a much smaller dataset, the training process lasts for 600 epochs with a similar hyper-paramter setup as Human3.6M. 

\subsection{Ablative Study}

\subsubsection*{Lie Algebra Pose Representation}

It has been shown in previous works that directly regressing the future poses with the ordinary "XYZ" pose representation \cite{AdvGeoAware, Shuang, Seq2seq-ZeroV} can easily produce unrealistic poses, we first validate this observation with our PoseMoNet by turning off the supervision at the Lie algebra pose representation outputted by the self-projection module. 
It is worth to highlight that, although the ordinary "XYZ" pose representation is most commonly used and there is no significant shortage of using it, we argue and empirically show that only the plain pose representation is insufficient in our work. We have 2 fundamental differences compared with existing works. First, to the best of our knowledge, a 3D MoCap pose sequence is required as the input for all the existing 3D future motion prediction method, where in our work, we only use 2D pose keypoint sequence as the only input to the task. Secondly, in our work, we are looking 1 more step further by considering a real world application scenario, we only leverage estimated 2D keypoint location without any human intervention, which make the problem more difficult and significantly different.
The result shows that both training and prediction are unstable and can easily lead to divergence. Therefore, we omit the quantitative comparison for this ablative experiment and show an expressive qualitative visualization.

\begin{figure}[t]
	\begin{center}
		\includegraphics[width=0.5\linewidth]{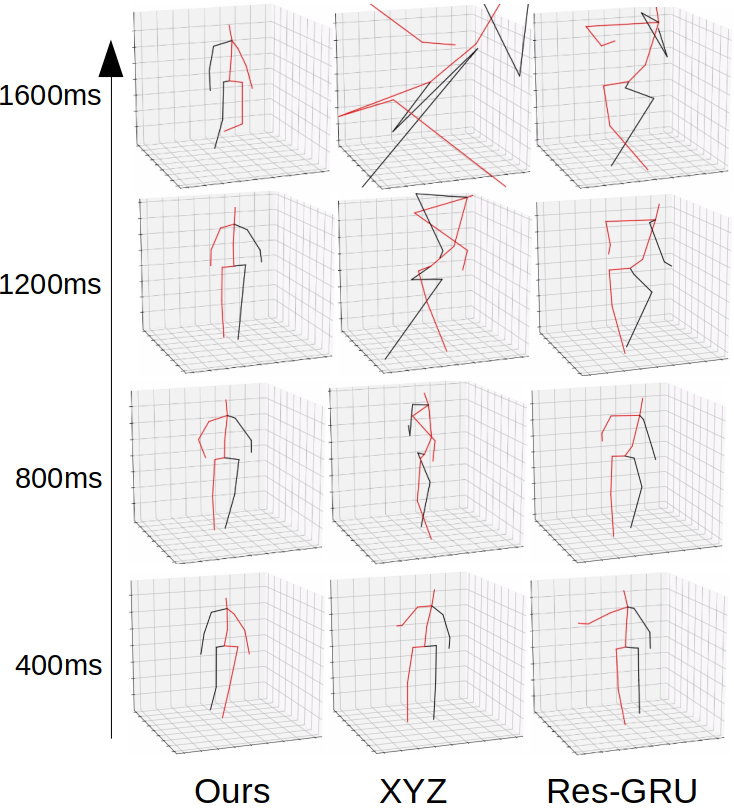}
	\end{center}
	\caption{A straightforward visualization for the ablative study of whether to include Lie algebra pose representation for predicting future motion. "XYZ" means the Lie algebra pose representation is not included in training under our framework. "Res-GRU" shows results directly obtained from the reproduced Res-GRU baseline.}
	\label{fig:h36m-motion}
\end{figure}

We demonstrate the visual results in Figure \ref{fig:h36m-motion}. As mentioned, we first train our framework without supervision on the Lie algebra pose representation. The result is shown in the middle column, denote as "XYZ". We also replicated the model proposed as Res-GRU \cite{Seq2seq-ZeroV} and show the result in the right column. We see that the previous competitive method may also suffer from the issue caused by no physical constraints. Finally, we add the results predicted by our framework trained completely with both pose representation and all loss terms. Not surprisingly, while other setups start to diverge on the predictions under a longer-than-usual context, the prediction from our framework, even after a 1600ms temporal context, is still natural and dynamic. This observation consolidates the effectiveness of including the Lie algebra pose representation in the training phase as well as using it to provide self-regularization with implicit physical constraints.

\subsubsection*{Multitask vs. Single-task}

We demonstrate the advantage of building the multitask framework over training two standalone models with ablative experiments. The first group of models is Pose Liter Networks (PLN) being trained without the Motion Generator Networks (MGN) followed after. Different input lengths are covered in our experiments: the sequences of 5, 9, 27, and 54 frames, are used to train 4 PLNs, respectively. We then train another group of 4 models with the exact same configuration but with the Motion Generator Network trained together. 
The comparison of the quantitative results between these two groups is shown in Table \ref{tab:multi}.

\begin{table}[t]
	\begin{center}
		\resizebox{0.5\textwidth}{!}
		{
			\begin{tabular}{@{}cccc@{}}
				\toprule
				& 9-frame     & 27-frame     & 54-frame     \\ \midrule
				w/o motion (GT) & 43.9 (31.4) & 41.5 (31.7) & 37.5 (27.1) \\
				w/ motion  (GT) & 42.3 (29.1) & 38.3 (27.5) & 33.4 (25.6) \\ \midrule
				w/o motion (SH) & 58.5 (45.7) & 53.2 (41.3) & 47.4 (28.5) \\
				w/ motion  (SH) & 56.8 (44.5) & 51.1 (40.9) & 45.3 (27.1) \\ \bottomrule
			\end{tabular}
		}
	\end{center}    
	\caption{Quantitative results of the ablative study on whether to train together with the motion generator network on Human3.6M dataset \cite{H36M}. We measure the performance of the model with both evaluation protocols on the test subjects. For each trained model, both MPJPE and P-MPJPE (in parentheses) are reported. GT means we use ground truth 2D keypoints during the training where SH denotes the prediction from a stacked hourglass model is leveraged.}
	\label{tab:multi}
\end{table}

\begin{figure*}[t]
	\begin{center}
		\includegraphics[width=1\linewidth]{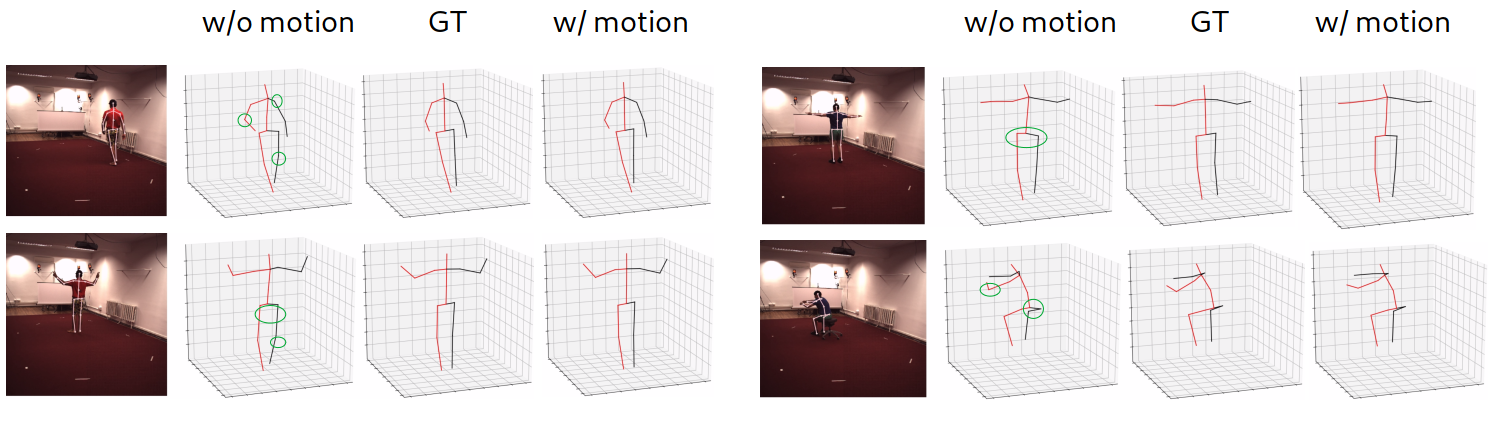}
	\end{center}
	\caption{Qualitative visualization for the ablative study w/o the Motion Generator Network (MGN). Green circles are added for helping to locate the failure cases visually in results from models trained without MGN.}
	\label{fig:h36m-qualitative}
\end{figure*}

The quantitative results provide us a better understanding of the benefits of introducing the multitask setting. We observe that, for varying lengths of input sequences, training the PLN along with the MGN leads to better quantitative results under both protocols. Surprisingly, if we focus on protocol 2, P-MPJPE,  the distance metric after the alignment at scale, rotation, and transformation, we notice a more significant boost over protocol 1 (MPJPE). We argue that by including the MGN, i.e., providing a broader temporal context in the future, our PLN obtained stronger constraints and resulted in predicting results that are closer to the ground truth, as shown in Figure \ref{fig:h36m-qualitative}.

For the sake of the completeness of the ablative study, we also performed experiments on removing the video pose estimation task, i.e., the supervision for 3D pose estimation is turned off and therefore only 2D pose sequence is used directly as input to generate future motion, and the training process suffered from huge fluctuation and diverged quickly, which consolidated the fact that 2D and 3D pose are not bijective therefore it can lead to massive ambiguity even in the latent space function mapping.

\subsubsection*{With vs. Without the Global Refinement Module}

\begin{table}[]
	\centering
	\resizebox{0.7\textwidth}{!}
	{
		\begin{tabular}{@{}ccccc@{}}
			\toprule
			& GRU & ConvED & GRU + ConvED w/o AW & GRU + ConvED w/ AW\\ \midrule
			MPJPE & 2.6 & 1.4 & 2.9 & 3.1 \\
			MAE & 0.04 & 0.07 & 0.05 & 0.05\\ \bottomrule
		\end{tabular}
	}
	\caption{Quantitative results of controlled experiments for testing the effectiveness of different types of refinement modules. All results are reported as the improvement obtained compared with no refinement. The MPJPE metric is used for evaluating the pose sequence estimation (the outputs from PLN). The Mean Angle Error is only employed to the first 200ms of generated future motion. AW stands for adaptive weighting.}
	\label{tab:refinement}
\end{table}

The proposed global refinement module consists of two baseline refinement networks mentioned previously. The effectiveness of the global refinement module is validated from two perspectives. To select the best architecture for pose trajectory refinement, we compared different settings for refining the entire pose trajectory under the following settings: using only the GRU-based model, using only the ConvED, and combining them to the proposed global refinement module. Quantitative results are obtained with different lengths of input video sequences and reported in Table \ref{tab:refinement}. We see that the GRU alone provides more improvement for pose estimation, which consolidated our observation that a significant amount of errors on pose sequence estimation is caused by the non-smooth prediction among consecutive frames. On the other hand, ConvED focuses more on the spatiotemporal context of the pose sequence, which not surprisingly enhanced the result of the motion generator more than the GRU. On combining results from both paths, the adaptive weighting mechanism has been shown to be helpful in terms of quantitative performance. 

Besides, the effectiveness of the refinement module along with the multitask vs. single task setting is also examined by first attaching the GR module only on the estimated 3D pose sequence, i.e., the output from the Pose Lifting Network, then we apply the GR module on the entire estimated pose sequence (predicted 3D pose + predicted future motion). Qualitative results is briefly reported in Table \ref{tab:refinement}.

\subsection{Evaluation on Human3.6M Dataset}

\subsubsection*{3D Pose Estimation in Video}

\begin{table*}[]
    \hspace{-3.6em}
	\resizebox{1.2\textwidth}{!}
	{
\begin{tabular}{@{}cccccccccccccccc|c@{}}
    \toprule
    MPJPE                                             & Direct. & Discuss & Eating & Greet & Phone & Photo & Pose  & Purch. & Sitting & SitD.     & Smoke & Wait  & WalkD. & Walk & WalkT. & Avg.  \\ \midrule
    Ionescu et al. $\star$ ~\cite{H36M}                      & 132.7   & 183.6   & 132.3  & 164.4 & 162.1 & 205.9 & 150.6 & 171.3  & 151.6   & 243.0     & 162.1 & 170.7 & 177.7  & 96.6 & 127.9  & 162.1 \\
    Martinez et al. $\star$ ~\cite{SimpleYet}                & 51.8    & 56.2    & 58.1   & 59.0  & 69.5  & 78.4  & 55.2  & 58.1   & 74.0    & 94.6      & 62.3  & 59.1  & 65.1   & 49.5 & 52.4   & 62.9  \\
    Hossain \& Little $\dagger$ ~\cite{ExploitingTemporal}   & 48.4    & 50.7    & 57.2   & 55.2  & 63.1  & 72.6  & 53.0  & 51.7   & 66.1    & 80.9      & 59.0  & 57.3  & 62.4   & 46.6 & 49.6   & 58.3  \\
    Zhao el al. $\star$ ~\cite{SemGCN}                       & 47.3    & 60.7    & 51.4   & 60.5  & 61.1  & \textbf{49.9}  & 47.3  & 68.1   & 86.2    & \uline{55.0}      & 67.8  & 61.0  & \textbf{42.1}   & 60.6 & 45.3   & 57.6  \\ 
    Yang et al. $\star$ ~\cite{yang20183d}                   & 51.5    & 58.9    & 50.4   & 57.0  & 62.1  & 65.4  & 49.8  & 52.7   & 69.2    & 85.2      & 57.4  & 58.4  & 43.6   & 60.1 & 47.7   & 58.6  \\
    Fang et al. $\star$ ~\cite{PoseGrammar}             & 50.1    & 54.3    & 57.0   & 57.1  & 66.6  & 73.3  & 53.4  & 55.7   & 72.8    & 88.6      & 60.3  & 57.7  & 62.7   & 47.5 & 50.6   & 60.4  \\
    Cai el al. $\dagger$ ~\cite{ST-GCN-Pose}                 & \uline{44.6}    & 47.4    & 45.6   & 48.8  & 50.8  & 59.0  & 47.2  & \uline{43.9}   & 57.9    & 61.9      & 49.7  & 46.6  & 51.3   & \uline{37.1} & 39.4   & 48.8  \\ Pavllo et al. $\ddagger$ ~\cite{VideoPose3D}             & 45.2    & \uline{46.7}    & \uline{43.3}   & \uline{45.6}  & \uline{48.1}  & 55.1  & \textbf{44.6}  & 44.3   & \uline{57.3}    & 65.8      & \uline{47.1}  & \textbf{44.0}  & 49.0   & \textbf{32.8} & \textbf{33.9}   & 46.8  \\
    Cheng et al. ~\cite{Occlusion}                          & -       & -       & -      & -     & -     & -     & -     & -      & -       & -         & -     & -     & -      & -    & -      & \uline{44.8}  \\

    Ours (9-frame)                                           & 54.3    & 60.3    & 51.9   & 55.9  & 58.4  & 66.2  & 59.9  & 55.6   & 65.7    & 68.2      & 56.4  & 58.8  & 57.6   & 48.4  & 50.0  & 56.8  \\
    Ours (27-frame)                                          & 46.0    & 51.4    & 44.3   & 49.1  & 53.6  & 59.3  & 47.5  & 46.9   & 65.5    & 75.7      & 49.6  & 50.5  & 52.4   & 43.7  & 46.2  & 51.1  \\
    Ours (54-frame)     & \textbf{42.7}    & \textbf{45.0}    & \textbf{40.5}   & \textbf{43.4}  & \textbf{46.4}  & \uline{51.4}  & \uline{46.0}  & \textbf{40.7}   & \textbf{52.3}    & \textbf{51.1}      & \textbf{44.2}  & \uline{44.1}  & \uline{43.4}   & 38.1  & \uline{38.3}  & \textbf{44.3}  \\ 
    
                                                             \midrule
    Chen et al. $\bullet$ ~\cite{Anatomy}                              & 41.4    & 43.5    & 40.1   & 42.9  & 46.6  & 51.9  & 41.7  & 42.3   & 53.9    & 60.2      & 45.4  & 41.7  & 46.0   & 31.5  & 32.7  & 44.1  \\
    Cheng et al. $\bullet$ ~\cite{Occlusion}                           & 36.2    & 38.1    & 42.7   & 35.9  & 38.2  & 45.7  & 36.8  & 42.0   & 45.9    & 51.3      & 41.8  & 41.5  & 43.8   & 33.1  & 28.6  & 40.1  \\
    \bottomrule
    \\[1cm]
    \bottomrule
    P-MPJPE                                           & Direct. & Discuss & Eating & Greet & Phone & Photo & Pose  & Purch. & Sitting & SitD.     & Smoke & Wait  & WalkD. & Walk & WalkT. & Avg.  \\ \midrule
    Martinez et al. $\star$ ~\cite{SimpleYet}                & 39.5    &43.2     &46.4    &47.0   &51.0   &56.0   &41.4   & 40.6   &56.5     &69.4       &49.2   &45.0   &49.5    &38.0   &43.1   &47.7 \\
    Hossain \& Little $\dagger$ ~\cite{ExploitingTemporal}   & 36.9    &37.9     &42.8    &40.3   &46.8   &46.7   &37.7   & 36.5   &48.9     &52.6       &45.6   &39.6   &43.5    &35.2   &38.5   &42.0  \\
    Yang et al. $\star$ ~\cite{yang20183d}                   & \textbf{26.9}    & \uline{30.9}     &36.3    &39.9   &43.9   &47.4   &\textbf{28.8}   &\textbf{29.4}    & \textbf{36.9}    &58.4       &41.5   &\textbf{30.5}   &\textbf{29.5}    &42.5   & 32.2  & 37.7 \\
    Cai el al. $\dagger$ ~\cite{ST-GCN-Pose}                 & 35.7    &37.8     &36.9    &40.7   &39.6   &45.2   & 37.4  & 34.5   &46.9     &\uline{50.1}       &40.5   & 36.1  &41.0    &29.6   & 33.2  & 39.0  \\
    Pavllo et al. $\ddagger$ ~\cite{VideoPose3D}             & 34.1    &36.1     & \uline{34.4}    &\textbf{37.2}   &\textbf{36.4}   &\uline{42.2}   &34.4   & 33.6   & 45.0    & 52.5      & \textbf{37.4}   &33.8   & 37.8   &\textbf{25.6}   &\textbf{27.3}   & 36.5 \\
    Cheng et al. $\bullet$ ~\cite{Occlusion}                         & -       & -       & -      & -     & -     & -     & -     & -      & -       & -         & -     & -     & -      & -     & -     & \textbf{34.1} \\
                                                             
    Ours (27-frame)                                          & 35.2    & 39.4    & 36.6   & 42.3  & 43.7  & 45.5  & 34.9  & 35.4   & 47.1    & 55.2      & 43.3  & 38.1  & 36.3   & 34.7  & 36.2  & 40.2 \\
    Ours (54-frame)                                          & \uline{28.4}    & \textbf{29.6}    & \textbf{33.9}   & \uline{38.5}  & \uline{37.4}  & \textbf{41.9}   & \uline{29.4}   & \uline{30.9}   &\uline{39.8}     & \textbf{49.7}      & \uline{38.5}  & \uline{31.6}   & \uline{31.8}   & \uline{28.2}  & \uline{31.7}  & \uline{34.7} \\ 
                                                             \midrule
    Chen et al. $\bullet$ ~\cite{Anatomy}                              & 32.6    &35.1     &32.8    &35.4   &36.3   &40.4   &32.4   &32.3    &42.7     &49.0       &36.8   &32.4   &36.0    &24.9   &26.5   &35.0 \\
    Cheng et al.$\bullet$  ~\cite{Occlusion}                           & 28.7    &30.3     &35.1    &31.6   &30.2   &36.8   &31.5   &29.3    &41.3     &45.9       &33.1   &34.0   &31.4    &26.1   &27.8   &32.8 \\
    \bottomrule
\end{tabular}
	}
	
	\caption{ Performance of 3D human pose estimation methods on the Human3.6M dataset, where our method is compared with the state-of-the-arts. Results evaluated under MPJPE (protocol \#1) and P-MPJPE (protocol \#2) are reported where \textbf{best in bold}, \uline{second-best underlined}. \\
	The first table focuses on MPJPE metric: Note that 2 results from \cite{Occlusion} are reported, the one with only the averaged result has the same problem setting with the compared literature, the one with complete results for each category involves augmented data that goes beyond the standard setting considered here. Methods leverage extra dataset or use extra data augmentations \cite{Anatomy, Occlusion} are included at the bottom of the table for the completeness purpose; they serve as a form of performance upper-bound for all the methods considered in our standard setting.  \\
	Moreover, the alternative metric of P-MPJPE is reported in the second table. Note here Cheng et al. \cite{Occlusion} has again two results being reported here, where the first one use the exact same configuration with other compared methods and the one noted with $\bullet$ used extra data augmentation step to improve the result.
    \\ $\star$: the corresponding method uses a single frame as input; \\$\dagger$: using multiple frames as input but with less or a similar number of frames compared to us; \\$\ddagger$: using more frames as input compared to us (243 vs. ours 54). \\ $\bullet$: using extra data augmentation to obtained better 2D keypoint sequence input or involved extra dataset in training
    }
    
	\label{tab:h36m-multitask}
\end{table*}

As shown in Table \ref{tab:h36m-multitask}, we compare with previous state-of-the-art methods for 3D pose estimation on the Human3.6M dataset. Different collections of methods are included. We have evaluated the methods proposed in \cite{SimpleYet, SemGCN} which focus on 3D pose estimation from a single image. For a fair comparison, we also compete with methods that leverage a video sequence as input \cite{ExploitingTemporal, VideoPose3D, ST-GCN-Pose, Anatomy, Occlusion, yang20183d}. It is important to note that a recently proposed work \cite{Occlusion} leverages extra data augmentation step to improve the 2D keypoint detection result, which further pushed the performance. Two results, one without the data augmentation step (which is using the same input compared with other methods) and one with their proposed data augmentation technique, are reported from \cite{Occlusion} and we include both results. Unfortunately, as the source code is not publicly available at the time we conduct experiments, we are not able to reproduce the proposed data augmentation. For the sake of completeness, we report both results in Table \ref{tab:h36m-multitask}.

We can see that compared with the state-of-the-art methods, by exploiting the temporal information along the sequence as well as the motion dynamics from future motion prediction task, our approach achieves the decent performance with the smallest error averaging across all actions and holding the 1st place for most categories of actions, such as Eating, Greet, Purch, etc. Interestingly, two methods \cite{ExploitingTemporal, VideoPose3D} that use very large context information ($\geq$ 243 frames) outperform our framework on action types "Walking" and "Walking Together". These two types of actions are highly predictable if being estimated from a larger context due to its cyclic nature, where the pose estimation task can be benefited. On the other hand, our method performs particularly well on more random actions, such as "Eating", "Photo", "Purch" and "Smoke" compared with existing methods. This observation meets the fact that a human does not need to see a relatively long presence of other people to estimate and understand their pose from 2D inputs.

Furthermore, our framework only needs to observe a relatively short sequence (54 frames) to produce sound predictions for both 3D pose estimation and motion prediction, which makes it comes with the potential to be deployed easier in a real-world real-time situation. We argue that the longer the context is, the more diminishing of returns, even potential negative effects, can be observed. This can be seen in the cases of estimating pose stochastic activities such as "Sitting", "Purch", "Posing" and "Discussing". Our method has more stable and quantitatively better results compared to methods that leverage a larger context \cite{VideoPose3D, Occlusion}. 
We have also noticed that existing works can produce better quantitative results with even a single-frame input \cite{VideoPose3D} though longer contexts are used in our framework, we argue that the architecture in \cite{VideoPose3D} at the single-frame setting requires significantly more computations as it is a 5-block residual architecture, also with more parameters, where our recurrent-based architecture can be more efficient. It is also worth to mention that the future motion predictions task suffer significantly from using an estimated 2D keypoint sequence as the original input, it also affects the performance of pose estimation, particularly when the sequence is short (i.e., 9-frame).

\begin{figure}[h]
	\begin{center}
		\includegraphics[width=0.95\linewidth]{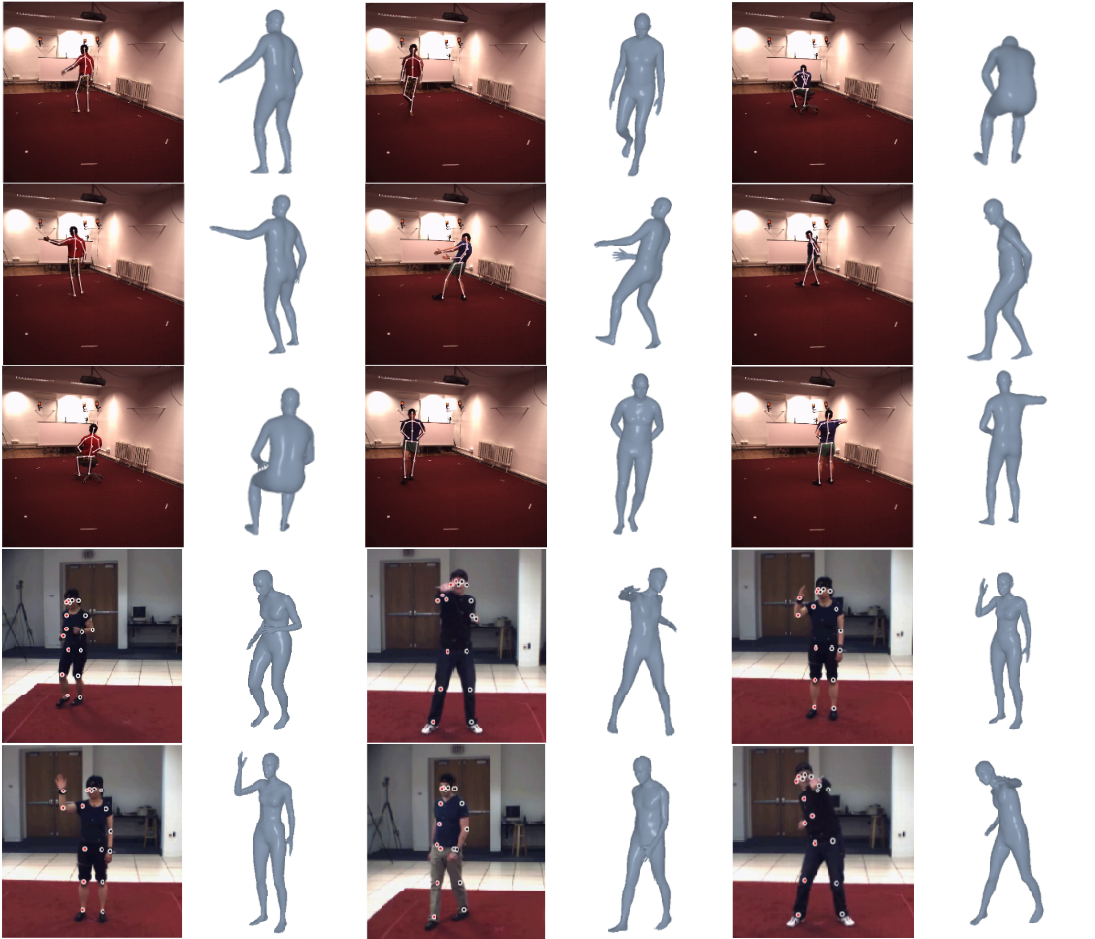}
	\end{center}
	\caption{The SMPL\cite{SMPL} shape-grounded visualization for estimated poses on Human3.6M and HumanEva-I. The shape visualization shows that our predicted poses are physically reasonable and there are is no body part cross each other. More visual results are included in the supplement video.}
	\label{fig:pose-shape}
\end{figure}

Consider that the skeleton visualization of the 3D poses is somehow abstract, we attempt to perform a reverse fitting process with the SMPL \cite{SMPL} shape model. A grounded visualization of the estimated poses can be obtained and is shown in Figure \ref{fig:pose-shape}. The predictions produced by our proposed framework can not only achieve decent quantitative performance but they are also physically reasonable that rarely the body parts have collision or cross each other. Furthermore, we also provide visualization for HumanEva-I here to demonstrate the generalization ability of the proposed work.

\subsubsection*{Future Motion Prediction}

\begin{table*}[]
    \hspace{-1.8em}
	\resizebox{1.1\textwidth}{!}
	{
        \begin{tabular}{|l|c|c|c|c|c|c|c|c|c|c|c|c|c|c|c|c|}
        	\hline
        	\multirow{2}{*}{Methods} & \multicolumn{8}{c|}{Discussion} & \multicolumn{8}{c|}{Greeting} \\ \cline{2-17}
        	& 80ms & 160ms & 320ms & 400ms & 560ms & 640ms & 720ms & 1000ms & 80ms & 160ms & 320ms & 400ms & 560ms & 640ms & 720ms & 1000ms \\ \hline
        	ERD \cite{ERD}                     & 2.22 & 2.38 & 2.58 & 2.69 & 2.89 & 2.93 & 2.94 & 3.11 & 1.70 & 2.04 & 2.60 & 2.81 & 3.29 & 3.47 & 3.55 & 3.43 \\ \hline
        	LSTM-3LR \cite{ERD}                & 1.80 & 2.00 & 2.13 & 2.13 & 2.29 & 2.32 & 2.36 & 2.44 & 0.93 & 1.51 & 2.27 & 2.54 & 2.97 & 3.05 & 3.12 & 3.09 \\ \hline
        	S-RNN \cite{S-RNN}                 & 1.16 & 1.40 & 1.75 & 1.85 & 2.06 & 2.07 & 2.08 & 2.19 & 1.33 & 1.60 & 1.83 & 1.98 & 2.27 & 2.28 & 2.30 & 2.31 \\ \hline
        	Res-GRU \cite{Seq2seq-ZeroV}       & 0.31 & 0.69 & 1.03 & 1.12 & 1.52 & 1.61 & 1.70 & 1.87 & 0.52 & 0.86 & 1.30 & 1.47 & 1.78 & 1.75 & 1.82 & 1.96 \\ \hline
        	Zero-velocity \cite{Seq2seq-ZeroV} & 0.31 & 0.67 & 0.97 & 1.04 & 1.41 & 1.56 & 1.71 & 1.96 & 0.54 & 0.89 & 1.30 & 1.49 & 1.79 & 1.74 & 1.77 & 1.80 \\ \hline
        	MHU \cite{MHU}                     & 0.31 & 0.66 & 0.93 & 1.00 & 1.37 & 1.51 & 1.66 & 1.88 & 0.54 & 0.87 & 1.27 & 1.45 & 1.75 & 1.71 & 1.74 & 1.87 \\ \hline
        	HMR \cite{Shuang}                  & 0.29 & 0.55 & 0.83 & 0.94 & 1.35 & 1.49 & 1.61 & 1.72 & 0.52 & 0.85 & 1.25 & 1.40 & 1.65 & 1.62 & 1.67 & 1.73 \\ \hline
        	AGED \cite{AdvGeoAware}            & 0.27 & 0.56 & 0.76 & 0.83 & 1.25 & -    & -    & 1.55 & 0.56 & 0.81 & 1.30 & 1.46 & -    & -    & -    & 1.69 \\ \hline
        	Traj \cite{TrajLearn}			   & 0.20 & 0.51 & 0.77 & 0.85 & 1.33 & -    & -    & 1.70 & 0.36 & 0.60 & 0.95 & 1.13 & -    & -    & -    & -    \\ \hline 
        	Ours                               & 0.30 & \textbf{0.55} & \textbf{0.82} & \textbf{0.97} & 1.38 & \textbf{1.55} & \textbf{1.65} & \textbf{1.85} & \textbf{0.53} & \textbf{0.78} & \textbf{1.28} & \textbf{1.41} & \textbf{1.69} & \textbf{1.67} & \textbf{1.71} & \textbf{1.79} \\ \hline
        	\multirow{2}{*}{Methods} & \multicolumn{8}{c|}{Posing} & \multicolumn{8}{c|}{Walking Dog} \\ \cline{2-17}
        	& 80ms & 160ms & 320ms & 400ms & 560ms & 640ms & 720ms & 1000ms & 80ms & 160ms & 320ms & 400ms & 560ms & 640ms & 720ms & 1000ms \\ \hline
        	ERD \cite{ERD}                     & 2.42 & 2.77 & 3.26 & 3.39 & 3.43 & 3.42 & 3.45 & 3.87 & 1.58 & 1.78 & 2.02 & 2.10 & 2.31 & 2.37 & 2.48 & 2.60 \\ \hline
        	LSTM-3LR \cite{ERD}                & 1.22 & 1.89 & 3.02 & 3.53 & 4.25 & 4.57 & 4.83 & 4.60 & 0.76 & 1.29 & 1.91 & 2.18 & 2.72 & 3.01 & 3.30 & 3.78 \\ \hline
        	S-RNN \cite{S-RNN}                 & 1.74 & 1.89 & 2.23 & 2.43 & 2.67 & 2.73 & 2.79 & 3.42 & 1.57 & 1.73 & 1.93 & 1.96 & 2.13 & 2.17 & 2.23 & 2.20 \\ \hline
        	Res-GRU \cite{Seq2seq-ZeroV}       & 0.41 & 0.84 & 1.53 & 1.81 & 2.06 & 2.21 & 2.24 & 2.53 & 0.56 & 0.95 & 1.33 & 1.48 & 1.78 & 1.81 & 1.88 & 1.96 \\ \hline
        	Zero-velocity \cite{Seq2seq-ZeroV} & 0.28 & 0.57 & 1.13 & 1.38 & 1.81 & 2.14 & 2.23 & 2.78 & 0.60 & 0.98 & 1.36 & 1.50 & 1.74 & 1.80 & 1.87 & 1.96 \\ \hline
        	MHU \cite{MHU}                     & 0.33 & 0.64 & 1.22 & 1.47 & 1.82 & 2.11 & 2.17 & 2.51 & 0.56 & 0.88 & 1.21 & 1.37 & 1.67 & 1.72 & 1.81 & 1.90 \\ \hline
        	HMR \cite{Shuang} 				   & 0.24 & 0.53 & 1.12 & 1.42 & 1.75 & 1.89 & 2.02 & 2.50 & 0.55 & 0.87 & 1.20 & 1.36 & 1.65 & 1.70 & 1.77 & 1.84 \\ \hline
        	AGED \cite{AdvGeoAware}            & 0.31 & 0.58 & 1.12 & 1.34 & -    & -    & -    & 2.65 & 0.50 & 0.81 & 1.15 & 1.27 & -    & -    & -    & 1.92 \\ \hline
        	Traj \cite{TrajLearn}			   & 0.19 & 0.44 & 1.01 & 1.24 & -    & -    & -    & -    & 0.46 & 0.79 & 1.12 & 1.29 & -    & -    & -    & -    \\ \hline 			
        	Ours                               & \textbf{0.24} & \textbf{0.57} & \textbf{1.15} & 1.45 & \textbf{1.80} & \textbf{2.10} & \textbf{2.15} & 2.67 & \textbf{0.53} & 0.88 & \textbf{1.15} & 1.41 & \textbf{1.70} & \textbf{1.76} & \textbf{1.83} & 1.99 \\ \hline
        \end{tabular}
	}
	\caption{The quantitative evaluation of that task of motion prediction and comparison with previous methods over 4 different action types on the H3.6m dataset. We compute Mean Angle Error (MAE) for quantitative evaluation. Note \textit{the comparison methods all rely on past 3D \textbf{ground-truth} pose sequence as input}, where our PoseMoNet \textbf{only uses} own estimated partial pose sequence, and relatively noisy 3D pose sequence as the input. To counter this important disadvantage, we choose to highlight as good results of our PoseMoNet that are in the top-3; This demonstrates our PoseMoNet could achieve comparable results even in an unfair situation.}
	\label{tab:h36m-motion}
\end{table*}

It is worth highlighting that our proposed work has a fundamental difference with the existing motion prediction literature where makes it is infeasible to have a completely fair and controlled experiment: we do not use the historical 3D ground truth as the input of the motion prediction task. In Table \ref{tab:h36m-motion}, all the previous state-of-the-art methods use past 3D ground truth as the context as well as a ground truth seed pose for generating the future poses. We highlight our results with bold font whenever our results can achieve the top-3 overall ranking. In spite of this, it is still clear that our quantitative results are comparable with methods that leveraged 3D ground truth. This helps to demonstrate the further potential of our proposed framework.

Following the literature, we evaluate the future motion generation task on four commonly used test action classes, namely, discussion, greeting, posing, and walking dog. Mean Angle Error (MAE) is reported for this task. We compare with milestone works in the literature including ERD \cite{ERD}, S-RNN \cite{S-RNN}, Res-GRU \cite{Seq2seq-ZeroV}, HMU \cite{LongTermMotion} and a recently proposed Lie algebra-based method, HMR \cite{Shuang}. Unfortunately, as mentioned before, there is no easy way to waive the disadvantage of not using 3D ground truth but we are still able to obtain comparable quantitative results.

\begin{figure}[]
	\begin{center}
		\includegraphics[width=1\linewidth]{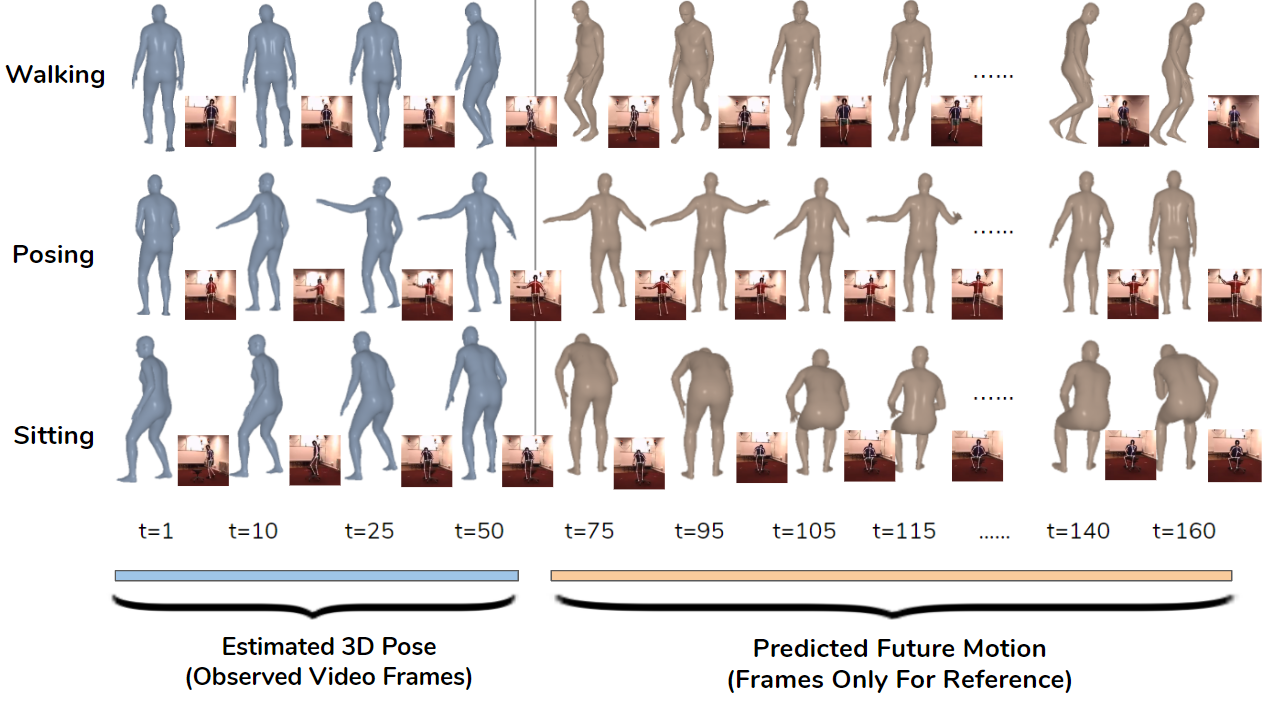}
	\end{center}
	\caption{Exemplar future motion prediction results. 
	Given as input a partial sequence of RGB images (shown in the blue \textit{present} temporal segment), we are to estimate poses from these observed images, and more importantly, to predict their future poses in the orange-colored \textit{future} time segment.
	Here, each row displays one predicted motion sequence: blue shapes are the partial sequence of estimated poses; orange shapes are the predicted future poses over time. Together they form a complete sequence of \textit{estimated} and \textit{predicted} poses. The unit of X-axis is frame number.}
	\label{fig:motion-shape}
\end{figure}

From Table \ref{tab:h36m-motion}, we can see that we can achieve comparable quantitative performance with the state-of-the-art methods \cite{AdvGeoAware, TrajLearn} within a 560ms time interval. Interestingly, the quantitative advantage of our method starts to deviate after 720ms. First, we believe this observation is acceptable, as there is, indeed, a potential drawback of using only the 2D pose sequence as the input. Besides, we argue that there are always trade-offs between the quantitative result and the quality of generated future motion as it has been pointed out that a very simple baseline, Zero-velocity \cite{Seq2seq-ZeroV}, which always produces the static mean pose outperforms many existing works quantitatively. 

We further validate the above observation by a qualitative visualization provided in \ref{fig:motion-shape}. Surprisingly, we found that the relatively high error measured under the MAE metrics not necessarily translates to a "bad" prediction of the future motion due to the fact that MAE is ambiguous which leads to the result of different sets of angles that may yield similar poses.
The visualization further consolidated that we are still able to generate sound poses for future motions. In particular, many previous methods have the problem that the predicted future motion starts to either drift or converge to a static pose \cite{ERD, LongTermMotion} but in our case, for example, the walking action can keep being dynamic after a relatively long period.

\begin{table*}[]
	\resizebox{1\textwidth}{!}
	{
        \begin{tabular}{|l|c|c|c|c|c|c|c|c|}
        	\hline
        	\multirow{2}{*}{Methods} & \multicolumn{4}{c|}{Discussion} & \multicolumn{4}{c|}{Greeting} \\ \cline{2-9}
        	& 80ms & 160ms & 320ms & 400ms  & 80ms & 160ms & 320ms & 400m \\ \hline
        	Zero-velocity \cite{Seq2seq-ZeroV} & 0.48          & 0.79    & 1.10    & 1.24    & 0.68     & 1.01    & 1.47 & 1.61 \\ 
        	AGED \cite{AdvGeoAware}            & 0.40          & 0.68    & 0.85    & \uline{0.97} & 0.62     & 0.92    & 1.30 & 1.42 \\ 
        	Traj \cite{TrajLearn}			   & \textbf{0.27} & \uline{0.56} & \uline{0.84} & \textbf{0.95} & \uline{0.55} & \uline{0.89} & \textbf{1.20} & \textbf{1.36} \\ \hline 
        	Ours                               & \uline{0.30}  & \textbf{0.55} & \textbf{0.82} & \uline{0.97} & \textbf{0.53}  & \textbf{0.78} & \uline{1.28} & \uline{1.41} \\ \hline
        	\multirow{2}{*}{Methods} & \multicolumn{4}{c|}{Posing} & \multicolumn{4}{c|}{Walking Dog} \\ \cline{2-9}
        	& 80ms & 160ms & 320ms & 400ms  & 80ms & 160ms & 320ms & 400m \\ \hline
        	Zero-velocity \cite{Seq2seq-ZeroV} & 0.45 & 0.70 & 1.35 & 1.57 & 0.64 & 1.10 & 1.56 & 1.73 \\ 
        	AGED \cite{AdvGeoAware}            & 0.39 & 0.63 & 1.31 & 1.62 & 0.65 & 1.01 & 1.33 & 1.54 \\ 
        	Traj \cite{TrajLearn}			   & \uline{0.26} & \textbf{0.56} & \uline{1.21} & \textbf{1.45} & \uline{0.59} & \uline{0.91} & \uline{1.20} & \uline{1.47} \\ \hline 			
        	Ours                               & \textbf{0.24} & \uline{0.57} & \textbf{1.15} & \uline{1.46} & \textbf{0.53} & \textbf{0.88} & \textbf{1.15} & \textbf{1.41} \\ \hline
        \end{tabular}
	}
	\caption{Quantitative results when training previous methods (by re-implementation or public code adoption) with estimated 3D pose sequence outputted by our system. We still use MAE as the metric and the results are calculated using the ground truth. Best in bold and second-best underlined.}
	\label{tab:h36m-motion-comp}
\end{table*}

Note that the shape visualization is only for a demonstrative purpose, we do not estimate the shape parameters in our framework. This fact also proves the effectiveness of our framework from another aspect where the predicted future motions are realistic and dynamic even we do not explicitly model the shape and constraints provided with the shape fitting process.

For pursuing a relative fair comparison, with the lighted shined by state-of-the-art methods, we have also either re-implemented \cite{Seq2seq-ZeroV, AdvGeoAware} or adopted public code base\cite{TrajLearn} to retrain previous methods by feeding them the estimated 3D historical pose sequence as the input. Due to the fact that different skeletal structures are used (different number of joints) as well as the input is no longer historical 3D ground truth poses, the quantitative results shacked significantly and we reported them in Table \ref{tab:h36m-motion-comp}. Clearly, the performance of the previous methods is significantly affected by the noisier input and our method is achieving the best or the second-best position among most of the time horizons. 

\subsection{Runtime and Efficiency}

A complete theoretical complexity analysis of our proposed framework is beyond the scope of this work. Therefore, we use a recently configured workstation to empirically evaluate the computational complexity. The workstation comes with an Intel i9-7900X CPU, RTX 2080 Ti GPU with 12 GB graphic memory, and 128 GB random access memory (RAM). We randomly select 1000 input 2D keypoint sequences of length 27, which is the same as our previous experiment, and use them to perform inference with different lengths of future motion conditions. The experiment covers both cases where the Lie-based pose representation is included or not and tests the runtime differences by involving the motion generator. Quantitative results is shown in Table \ref{tab:runtime}. Using recurrent architecture in our framework leads to a step-by-step inference nature, i.e., for inference at time step $t+1$, the computation at time step $t$ must be finished, the computational cost of the proposed method grows at a linear scale, therefore, for both the training and inference stage, the inference and training time only depend on how many steps we have for pose estimation and future motion generation. Real-time applications then become fully considerable. 
The recurrent architecture can be a better fit when the input sequence is relatively short (i.e., with 9, 27, 54-frame setting), however, convolutional architectures \cite{VideoPose3D} can be more efficient when the longer input sequence is involved (i.e., over 100-frame input). From a potential application aspect, it should be decided on the application end which one is a better fit.

\textcolor{blue}{
\begin{table}[]
\centering
\begin{tabular}{@{}cccc@{}}
\toprule
\# of Future Time Steps      & 0    & 27    & 54    \\ \midrule
w\textbackslash Lie    & 83ms & 140ms & 189ms \\
w\textbackslash{}o Lie & 57ms & 113ms & 157ms \\ \bottomrule
\end{tabular}
\caption{An empirical evaluation of the computational complexity. Given 2D pose sequences of length 27, we try to inference a different length of future motion. In the table, different comparison conditions are included. When the number of future motion time step is 0, it indicates we only tested the framework with the pose estimation network (i.e., generating 0 future motion). We also compare the affects by involving the Lie-based pose representation.}
\label{tab:runtime}
\end{table}
}


\section{Conclusion}
\label{sec:conclusion}

The closely related, but often separately considered tasks of 3D human pose estimation and future motion prediction are jointly tackled in this paper. By doing so, a dedicated approach, PoseMoNet, is developed to exploit the innate connections of both tasks: 
The proposed framework, to the best of our knowledge, is the first work to tackle the 3D pose sequence estimation and future motion prediction together.
A novel mixture-of-expert self-projection module is introduced to implicitly leverage the physical constraints provided by the Lie algebra pose representation; 
Multiple granularities of multitask settings are investigated and leveraged in the proposed framework. 
We demonstrated that 3D MoCap data is not a sole requirement for generating future 3D motion by combining it with the 3D pose sequence estimation task.
Empirical experiments on well-known Human3.6M and HumanEva-I benchmarks demonstrate the competitive performance in addressing the problems of 3D pose sequence estimation and future motion generation. 
Future work may focus on multiple aspects. An interesting direction is that an investigation beyond the estimation of current and future single poses, including estimation of current and future 3D shapes, and further investigation into visual scenarios involving the interactions of multiple people. In addition, recurrent architectures are the major components in our proposed work, future works may pay attention to other advanced research efforts such as transformer architectures that can avoid step-by-step inference caused by the recurrent architecture as well as graph convolution networks that can leverage the skeletal property of articulated objects.

\bibliography{egbib.bib}
\end{document}